\documentclass[runningheads]{llncs}
\usepackage[T1]{fontenc}
\usepackage{bm}
\usepackage{xcolor}
\usepackage[bottom]{footmisc}

\usepackage{caption}
\usepackage{acronym}
\usepackage{cite}
\usepackage{balance}
\usepackage{multicol}
\usepackage{amsmath,amssymb,amsfonts}
\usepackage{algorithmic}
\usepackage{graphicx} % Required for including images
\usepackage{hyperref} 
\usepackage[pdf]{graphviz}
\usepackage{booktabs}
\usepackage{subfig}
\usepackage{comment}

\usepackage{tikz}
\usepackage[table]{xcolor}
\usetikzlibrary{shapes.geometric}
\usetikzlibrary{positioning}
\usetikzlibrary{shapes,arrows,positioning}
\usepackage{amsmath}
\usepackage{textcomp}
\usepackage{xcolor}
\usepackage{hyperref}
\usepackage{graphicx}
\begin{document}
%
% \title{Hybrid guided variational autoencoder for rapid visual place recognition\thanks{Supported by organization x.}}
\title{Hybrid guided variational autoencoder for visual place recognition}

% \author{Ni Wang\inst{1}\orcidID{0000-1111-2222-3333} \and
% Zihan You\inst{2}\orcidID{1111-2222-3333-4444} \and
% Emre Neftci\inst{3}\orcidID{2222--3333-4444-5555} \and
% Thorben Schoepe\inst{4}\orcidID{2222--3333-4444-5555}}
\author{Ni Wang\inst{1} \and
Zihan You\inst{2} \and
Emre Neftci\inst{3} \and
Thorben Schoepe\inst{4} }
\authorrunning{N. Wang et al.}
\institute{Amazon Development Center Germany GmbH, Berlin, Germany \and
Southeast University, Nanjing, China \and
Forschungszentrum Jülich GmbH, Aachen, Germany\and
imec, Luven, Belgium
}
% \institute{Paris-Saclay University, Orsay, France \and
\acrodef{VPR}[VPR]{Visual Place Recognition}
\acrodef{VAE}[VAE]{Variational Auto Encoder}
\acrodef{ANN}[ANN]{Artifical Neural Network}
\acrodef{SNN}[SNN]{Spiking Neural Network}
\acrodef{ReLU}[ReLU]{Rectified Linear Unit}
\acrodef{BPTT}[BPTT]{back propagation through time}
\acrodef{STDP}[STDP]{spike time dependent plasticity}
\maketitle              % typeset the header of the contribution
\begin{abstract}
Autonomous agents such as cars, robots and drones need to precisely localize themselves in diverse environments, including in GPS-denied indoor environments. 
One approach for precise localization is visual place recognition (VPR), which estimates the place of an image based on previously seen places.
State-of-the-art VPR models require high amounts of memory, making them unwieldy for mobile deployment, while more compact models lack robustness and generalization capabilities. This work overcomes these limitations for robotics using a combination of event-based vision sensors and an event-based novel guided variational autoencoder (VAE). 
The encoder part of our model is based on a spiking neural network model which is compatible with power-efficient low latency neuromorphic hardware.
The VAE successfully disentangles the visual features of 16 distinct places in our new indoor VPR dataset with a classification performance comparable to other state-of-the-art approaches while, showing robust performance also under various illumination conditions. 
When tested with novel visual inputs from unknown scenes, our model can distinguish between these places, which demonstrates a high generalization capability by learning the essential features of location.
Our compact and robust guided VAE with generalization capabilities poses a promising model for visual place recognition that can significantly enhance mobile robot navigation in known and unknown indoor environments.

\end{abstract}

\section{Introduction}
\label{intro}
Enabling mobile robots to efficiently localize themselves within ambiguous spaces is fundamental to their safe and effective operation. 
Various approaches contribute to achieving this, such as path integration utilizing Inertial Measurement Units (IMU), scan matching with Lidar \cite{positioning}, and notably, \ac{VPR}. 
VPR is a task that aims to predict the place of an image (called query) based solely on its visual features \cite{whatis}.
VPR encompasses a crucial aspect of enabling robots to develop awareness of their surroundings, aiding in precise localization and mapping. 
A large variety of hand build approaches and also neural network approaches for \ac{VPR} have been developed in the last decades \cite{where}. Most of them are either computationally expensive or do not generalize well to unseen places. 
However, robotic agents require real-time capable, compact, low latency and low power algorithms in order to operate efficiently. Furthermore, a potential for generalization would enable their use in unknown environments without requiring expensive continual learning. Event-based cameras in combination with neuromorphic processors are a good fit for these hardware prerequisites. Event-based cameras sample log-intensity brightness changes on the single pixel level, leading to an asynchronous stream of information with high temporal resolution, low latency, and low power consumption, making event-based cameras a promising candidate for real time low power robotics \cite{eventcamera}. Neuromorphic processors are a new type of brain-inspired hardware which are optimized to emulate \acp{SNN} in real time, and can therefore directly process event-camera data in a massively parallel manner \cite{neuromrophicsurvey2022}. These processors are low power and low latency due to their sparse activation function and their massively parallel communication fabric. 

In this work we aim at a hardware friendly low-latency, low-power \ac{VPR} by developing a compact neural network algorithm that can utilize current event-based cameras and neuromorphic processors. We base our work on the \ac{VAE} model from \cite{beta}. It is a good fit due to its capability to disentangle important from unimportant features, its potential to identify and separate unknown input classes and its compact size with only 4.45M parameters and 917K neurons. The model is well suited for extreme edge robotics due to its event-based and partially spiking nature. The hybrid \ac{VAE} consists of a \ac{SNN} encoder and a classical \ac{ANN} decoder utilizing the \ac{ReLU} activation function. The \ac{SNN} encoder receives sensory input from an event-based camera. 
We record a new dataset for indoor event-camera \ac{VPR}. We train the guided \ac{VAE} on this new dataset, evaluate its performance and characterize its latent space representation while presented with known and unknown places. 
This is, to our knowledge, the first work that successfully applies a hybrid \ac{VAE} to the task of \ac{VPR}. By incorporating our model in a system of event based sensors and neuromorphic hardware we are aiming towards a new generation of brain inspired low-power, low-latency robotics.
The main contributions of this research project are:
\begin{itemize}
\item A full set of open-source event/RGB dataset `\textit{Aachen-indoor-VPR}' recorded with a mobile robot maneuvering within an office-like arena, which encompasses two FOVs and two levels of illumination, along with an additional dataset of robot maneuvers recorded in four new places.

\item Implementation and improvement of a hybrid guided VAE \cite{gesture} on the new task of VPR while exploring into a smaller latent space, resulting in a compact, low-power low-latency and robust indoor localization approach.   

\item Assessment on the capability of generalization and analysis into latent variable activity of this model.
\end{itemize}

The subsequent sections of this report firstly investigate related work in the domain of VPR and disentangled VAEs.
Section \ref{dataset} details the procedures of recording the dataset; Section \ref{method} explores two SNNs on VPR task; Section \ref{results} evaluates the experimental findings; Finally it summarizes insights from this work while planning future steps. 
Our code and dataset can be accessed at \cite{niart}.

\section{Related Work}
%This section reviews the state-of-the-art in \ac{VPR} and disentangled VAEs, including the challenge of distinguishing visually similar places, a key issue in VPR.
\paragraph{VPR Models} Traditional VPR methods using frame-based cameras often rely on CNN-based global descriptors, and comprehensive overviews can be found in recent surveys \cite{meet, tutorial}.
NetVLAD\cite{netvlad}, a prominent CNN architecture with VLAD-style aggregation, remains a widely adopted baseline for VPR across a variety of settings
\cite{ensembles, datasets2022yudin}.
%\cite{eventvlad2021, ensembles, arcanjo2024multitechnique, indoorvpr2023, datasets2022yudin}.
Despite its success, its large size (e.g., 88M parameters, 13M neurons with VGG16) makes it impractical for neuromorphic hardware, which supports currently 0.3 to 1M neurons and 120M parameters per chip \cite{speck, daviesloihi22021, truenorth2015}. Other CNN-based methods, like EigenPlaces \cite{whatis}, also face challenges due to high computational requirements and poor robustness in unseen environments.
Efforts like convolutional autoencoders \cite{cae} aim to reduce dimensionality and improve robustness.
In parallel, event-based cameras have become more popular as an alternative sensing modality. FE-Fusion-VPR \cite{conventional} fuses frames and events, aiming to improve robustness in challenging lighting and motion conditions. 
Event-VPR \cite{end} proposed an end-to-end weakly supervised network that directly processes event streams and reports improved performance over earlier ensemble-style event-VPR baselines \cite{ensembles}.
However, these deep event-based architectures typically remain complex, both in parameter count and in dense computation, and thus are still difficult to deploy efficiently on neuromorphic hardware.

Compact VPR approaches using event-based cameras and \ac{SNN}s has emerged as a promising research direction.
For example, ensemble-based event-camera place recognition under changing illumination aggregates multiple temporal windows to improve stability across day/night and other lighting shifts\cite{ensemble}.
In addition, Spike-EVPR\cite{spike} introduces an event-based spiking residual network with cross-representation aggregation, showing that SNN-style processing can be competitive when paired with suitable event representations.
Nevertheless, most of these work focus on outdoor scenarios.
Beyond pure event-stream learning, \cite{australia} introduced an SNN-based method with 600K parameters, demonstrating competitive VPR performance.
However, that approach relies on converted RGB imagery rather than native event streams and lacks evidence for cross-scene generalization.
%Other frame-based methods that compare incoming observations directly with stored traversals, such as SeqSLAM\cite{seq}, require significant memory, limiting their use in lightweight robots.

In summary, existing VPR methods span CNN- to SNN-based approaches, each exhibiting distinct advantages and limitations. 
Despite this progress, there remains a clear demand for hardware-efficient models with strong generalization capability.
Moreover, while prior work has introduced event-based datasets for outdoor VPR\cite{howmany, nyc} and RGB/LiDAR datasets for indoor VPR\cite{finland, nyc_indoor, datasets2022yudin}, open-source event-based datasets for indoor VPR remain scarce. 
This gap motivates the introduction of a new event-based indoor VPR dataset in this work.
\paragraph{Disentangled \ac{VAE}s}
Disentangled representation learning focuses on developing models that can identify and separate key features within observable data in representation form \cite{drl}. While \ac{VAE}s are commonly used to encode complex data into a latent space, they often mix various features together, making it difficult to isolate and manipulate specific traits \cite{disentangle}.

Recent efforts aim to tackle this issue by disentangling these features, allowing control over distinct aspects like spatial location, object identity, or lighting, while keeping the generative capabilities of traditional VAEs. This separation creates more interpretable and manageable latent spaces, enhancing our ability to understand and manipulate the data. Disentangled VAEs also improve generalization \cite{role} and transfer learning, helping models learn critical features independently and perform better across different datasets.
\cite{generalized} have discussed broad classes of VAE variants ($\beta$-VAE, $\sigma$-VAE, Dynamic-VAE, etc.) targeting improved disentanglement through inductive biases. 
\cite{2025variational} introduces DISCoVeR, a dual-latent VAE that explicitly separates shared vs. condition-specific factors with theoretical guarantees and improved disentanglement.
\cite{quantized} combines discrete latent quantization with total correlation regularization to enhance disentanglement quality and reconstruction. \cite{disentangle} proposed Guided-VAE by using an adversarial excitation and inhibition mechanism to encourage the disentanglement of the latent variables, which successfully disentangles factors like pose, identity, or style.
Inspired by that work, a hybrid Guided-VAE achieved 87\% accuracy in classifying hand gestures on the DVS Gesture dataset \cite{gesture}.

In this study, we aim to apply disentangled VAEs to capture location-related features, with the goal of cross-scene generalization without additional training.

%%%%%%%%%%%%%%%%%%%%%%%
\begin{figure}[htbp]
  \centering
  \begin{minipage}{.35\textwidth}
    \centering
    \includegraphics[height=6cm]{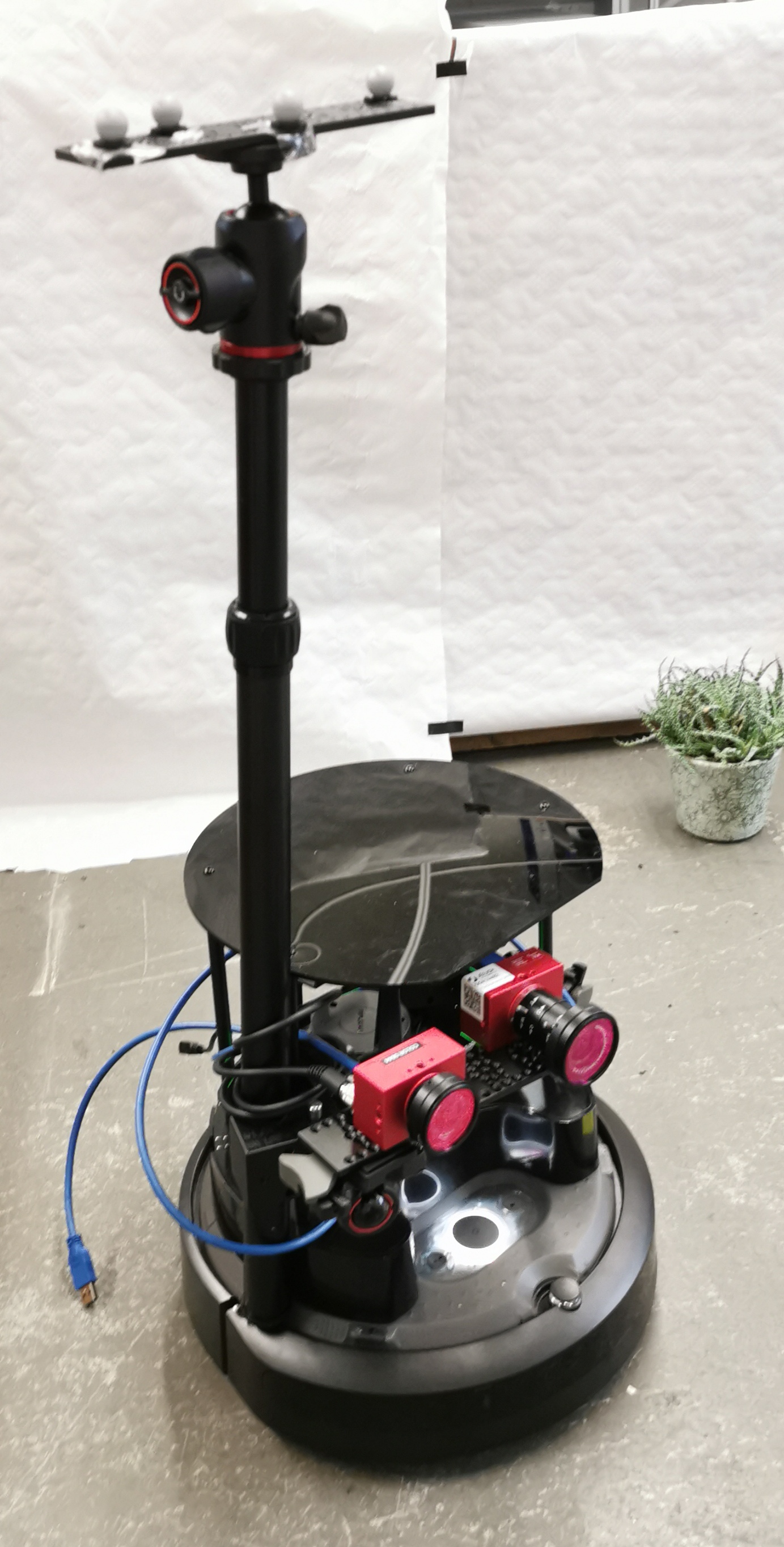}
  \caption{Turtlebot4 equipped with optical markers and a pair of event cameras.}
  \label{robot}
  \end{minipage}%
  \hspace{5pt}% Adjust the space as needed
  \begin{minipage}{.62\textwidth}
    \centering
    \includegraphics[height=4.5cm]{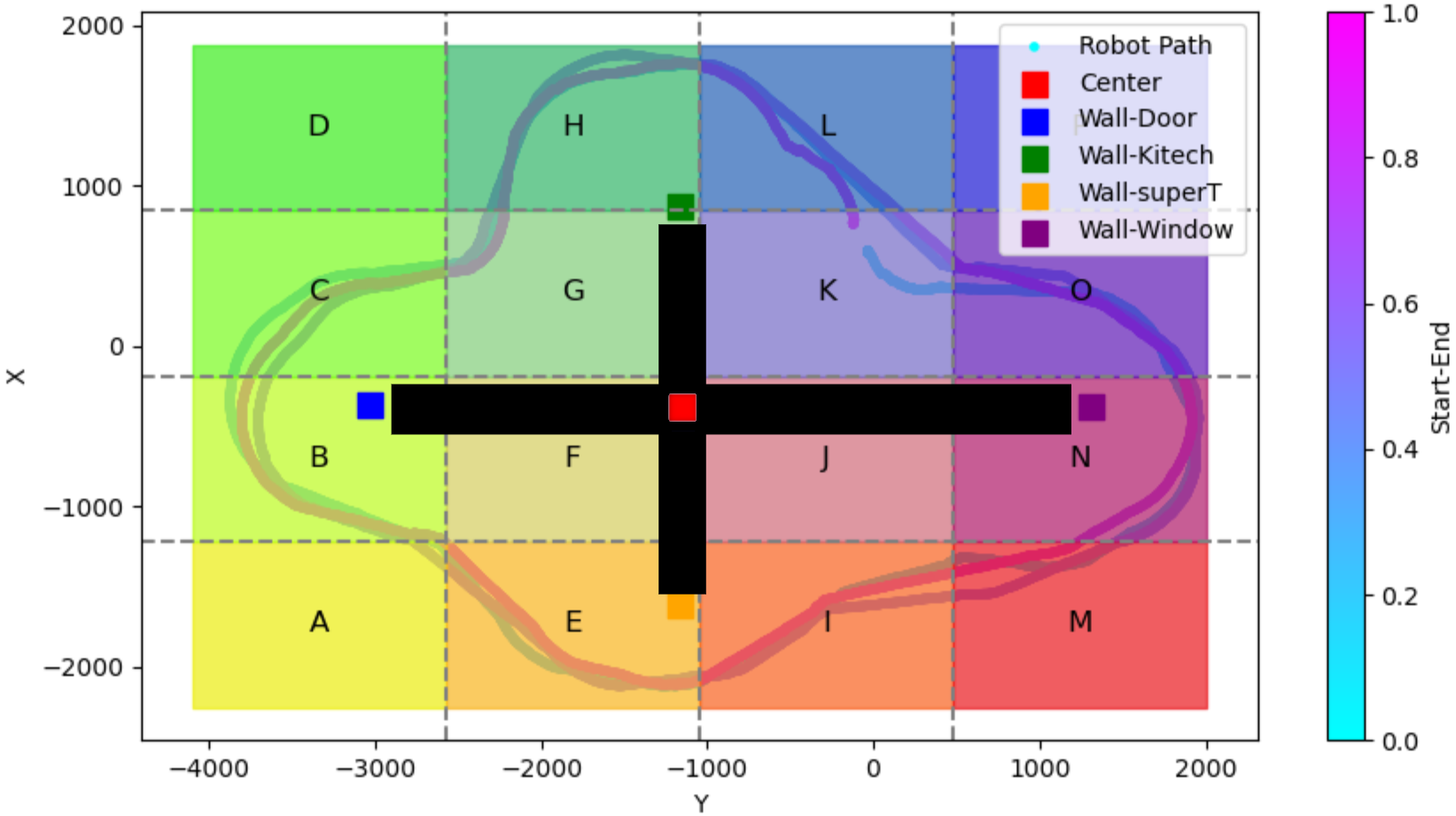}
 \caption{Layout of walls in the office-like environment (black `cross' bars) and the manoeuvre routine of robot during dataset recording (curve in color-map, counter-clockwise). The color of each cell denotes the coloring of each class in upcoming T-SNE visualization.}
  \label{arena}
  \end{minipage}
\end{figure}
\section{Dataset Collection \& Preprocessing}
\label{dataset}
This section provides a comprehensive overview of the acquisition and preprocessing of the event camera dataset used to train and test the proposed models.
\paragraph{Hardware setup \& Dataset recording}
\label{recording}
The dataset is collected using the Turtlebot4\cite{turtle} platform with two `DAVIS 346' event cameras mounted at the front (Figure~\ref{robot}). The left camera has a 12mm focal length, and the right has 2.5mm. Only data from the 2.5mm-camera is used.
The environment is a 6m$\times$4m artificial office-like space with four walls and various objects like tables, books, and bins (Figure~\ref{arena}). A motion capture system\cite{vicon} tracks the robot's location, providing labels for supervised learning and evaluation.
Three recordings were made by manually driving the robot along predefined routes. Two were under normal lighting, and one in dim conditions.

\begin{comment}
\begin{figure}[h]
\begin{center}
\includegraphics[height=.87in]{images/20231024_175824.jpg}
\includegraphics[height=.87in]{images/20231024_175836.jpg}
\includegraphics[height=.87in]{images/20231024_175849.jpg}
\includegraphics[height=.87in]{images/20231024_175530.jpg}
%\includegraphics[height=1.5in, angle =90]{images/20231024_175903.jpg}
\caption{An artificial office alike arena for dataset recording. Yellow stickers on the floor are predefined waypoints for manually driving the robot.}
\label{office}
\end{center}
\vskip -0.3in
\end{figure}
\end{comment}

\begin{figure}[htbp]
\begin{center}
\includegraphics[height=2.36cm]{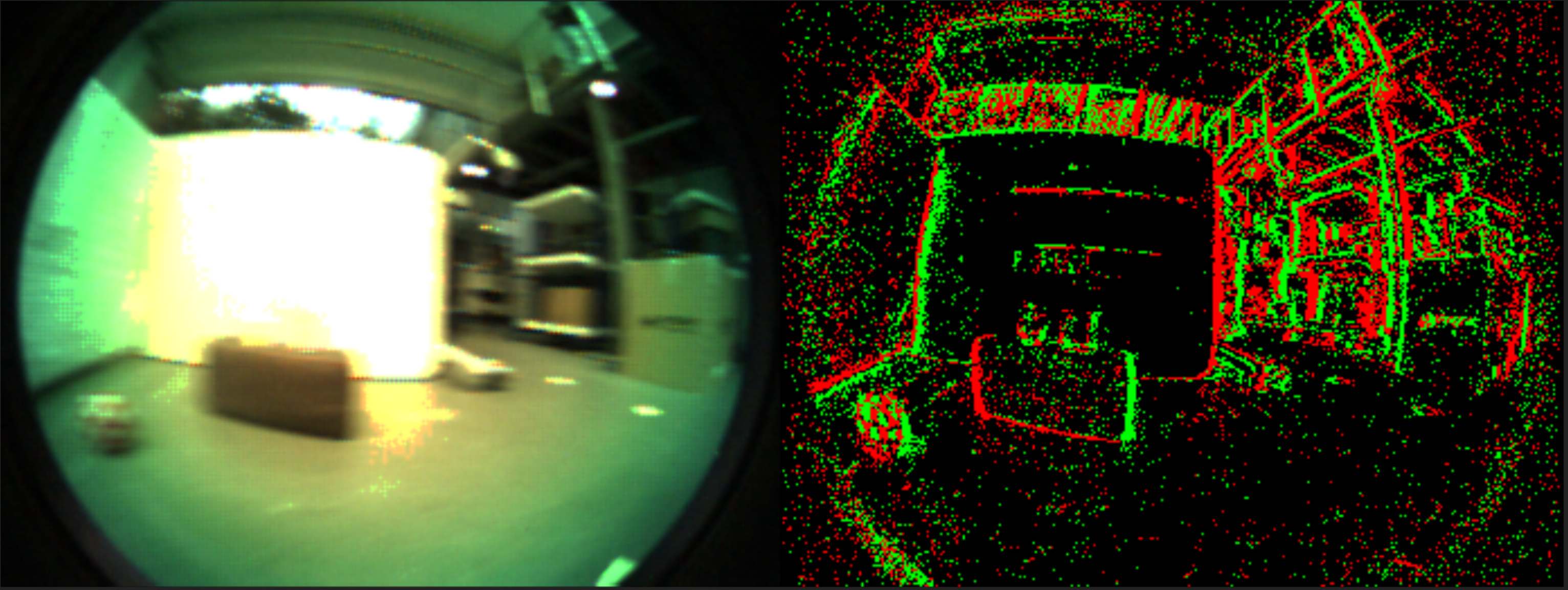}
\includegraphics[height=2.36cm]{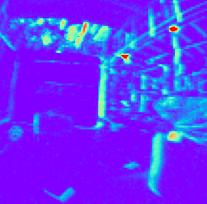}
\includegraphics[height=2.36cm]{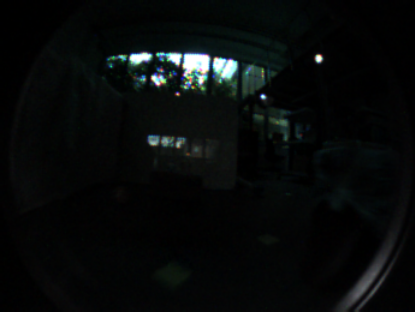}
%\caption{Typical RGB images from normal-illumination dataset (top and bottom-left), the rendered event frame (bottom-left 2nd), the event frame after preprocessing (bottom-left 3rd), and the RGB image of the same place in darkness (bottom-right).}
\caption{A typical RGB image from normal-illumination dataset (left), the rendered event frame (left 2nd), the event frame after preprocessing (left 3rd), and the RGB image of the same place in darkness (right).}
\label{rgb}
\end{center}
\end{figure}

\paragraph{Dataset preprocessing} Before training, the event stream is converted into 50 event frames per sample with a 2ms window and 128$\times$128 resolution. Figure~\ref{rgb} shows RGB and event data from the camera at different preprocessing stages. There are 1,500-1,700 event samples per recording. 
First recording is used for training, and second recording for testing. For varying lighting, first and third recordings are mixed and split evenly for training and testing.
%Table ~\ref{overview} summarized the recording, pre-processing and usage of the newly recorded datasets in this work. 

\section{Methodology}
\label{method}
This section describes how to transform high-dimensional event camera inputs into disentangled low-dimensional latent representations in order to extract spatial information. 
An unguided and a guided VAE from \cite{gesture} are investigated and modified to discover the influence of guidance on spatial latent representations. 

%In an effort to assess the models' performance across varying input characteristics, as well as enable the networks to capture both detailed and general features at the same time leading to a more precise representation, both networks undergo training and testing procedures using datasets captured by both wide and narrow FOV cameras.

We chose a hybrid model with an \ac{SNN} encoder and an \ac{ANN} decoder because of three reasons. First of all, our \acp{SNN} are trained by \ac{BPTT} which requires many more resources than back-propagating through a conventional \ac{ANN}. Therefore, a hybrid model requires less resources than a fully spiking model and might learn faster. Secondly, only the first few layers of a neural network trained by \ac{BPTT} are thought to carry highly relevant temporal information which can only be captured by an \ac{SNN} but not by a vanilla \ac{ANN} without recurrency. Hence, an \ac{SNN} is only required in the first layers. Third, only the encoder part will be implemented on neuromorphic hardware after training on a GPU.

\subsection{Hybrid unguided $\beta$-VAE}
In the hybrid unguided $\beta$-VAE, the SNN encoder comprises a sequence of four discrete leaky integrate and fire (LIF) convolutional layers, succeeded by linear layers, which output the latent state distribution (mean $\mathbf{\mu}$, variance $\sigma$) from which a vector $\mathbf{z}$ is sampled with dimensionality of 64 variables.
The network is complemented by a transposed convolutional network activated by rectified linear unit (ReLU) function serving as the decoder.
The architecture of this network is explained in Figure~\ref{guided}. The standard $\beta$-\ac{VAE} loss function is:

% a coefficient $\beta$ to scale the importance of the Kullback-Leibler (KL) divergence term in a normal VAE compared to the reconstruction term.  
%The overall loss function of this hybrid unguided VAE is thus:
{\footnotesize
\begin{equation}
\mathcal{L} = \mathbb{E}_{q_{\phi}(z \,|\, x)}\left[\log p_{\theta}(x \,|\, z)\right] - \beta \cdot \text{KL}\left(q_{\phi}(z \,|\, x) \,||\, p(z)\right),
\end{equation}
}
where 
%$\mathcal{L}$ represents the VAE loss, 
$\mathbb{E}_{q_{\phi}(z|x)}\left[\log p_{\theta}(x|z)\right]$ is the reconstruction loss, which measures how well the decoder $p_{\theta}(x|z)$ reconstructs the input data $x$ from the latent variables $z$, and $\text{KL}\left(q_{\phi}(z|x) || p(z)\right)$ 
%which denotes the KL divergence loss term between {\color{red} X and Y}, 
which denotes the KL divergence loss term between the encoder's output distribution over latent variables given the input $q_{\phi}(z|x)$ and the prior distribution over the latent variables $p(z)$, 
responsible for regularizing the latent variable distribution. The coefficient $\mathbf{\beta}$ controls the trade-off between these two components of the loss function, allowing for a flexible adjustment of the model's behavior.

\subsection{Hybrid guided VAE}
The second neural network extends the prior hybrid beta VAE by integrating a guidance mechanism after the encoder in a supervised manner.
In order to guide the VAE to encode location features, the entire arena is evenly divided into 4 $\times$ 4 = 16 labeled square cells as shown in Figure~\ref{arena}.
The decoder uses the entire latent vector $\mathbf{z}$ of 64 variables to reconstruct the event frame. 

\begin{figure}[htbp]
\begin{center}
\centerline{\includegraphics[width=1\textwidth]{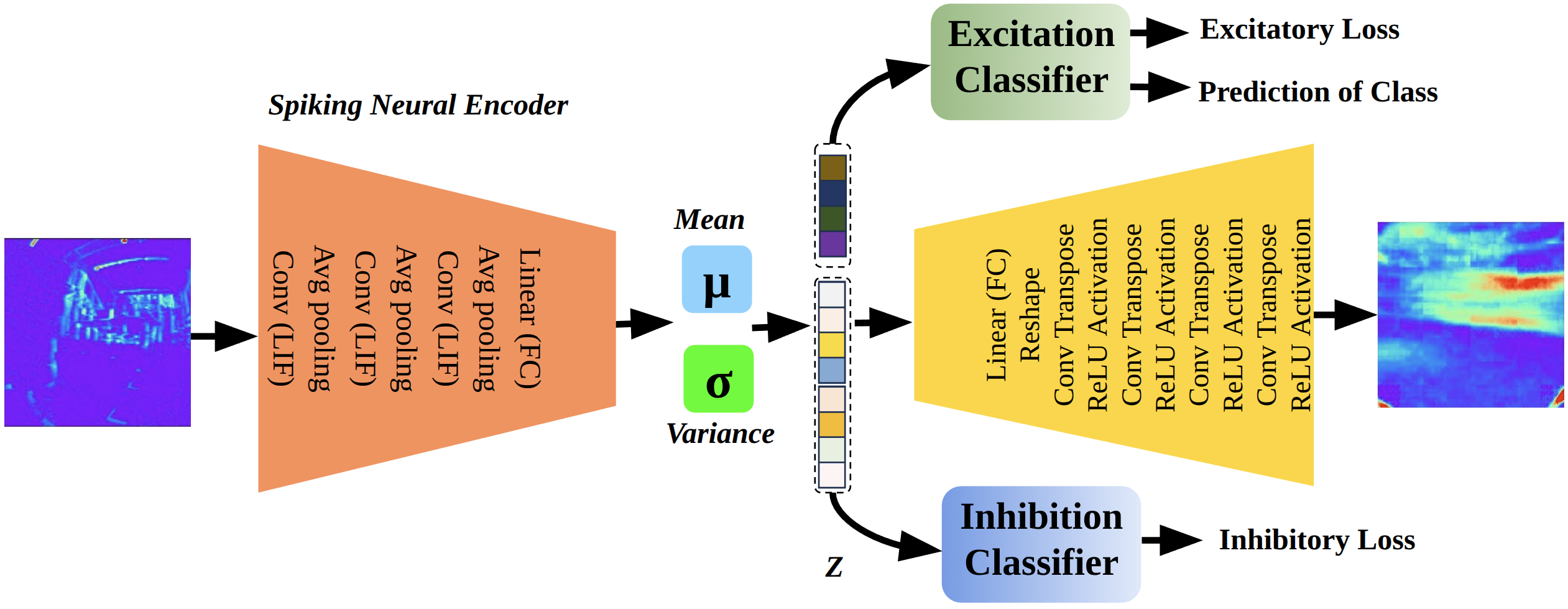}}
\caption{This VAE is unguided without the two additional classifiers. 
It is guided when jointly trained with an excitation classifier which takes the first 16 latent variables as input, and an inhibition classifier which takes the remaining as input.}
%The details of architecture of the hybrid guided VAE is summarised in Table~\ref{arct}.
\label{guided}
\end{center}
\vskip -0.2in
\end{figure}
\begin{figure}[t]
\begin{center}
\includegraphics[height=0.99in]{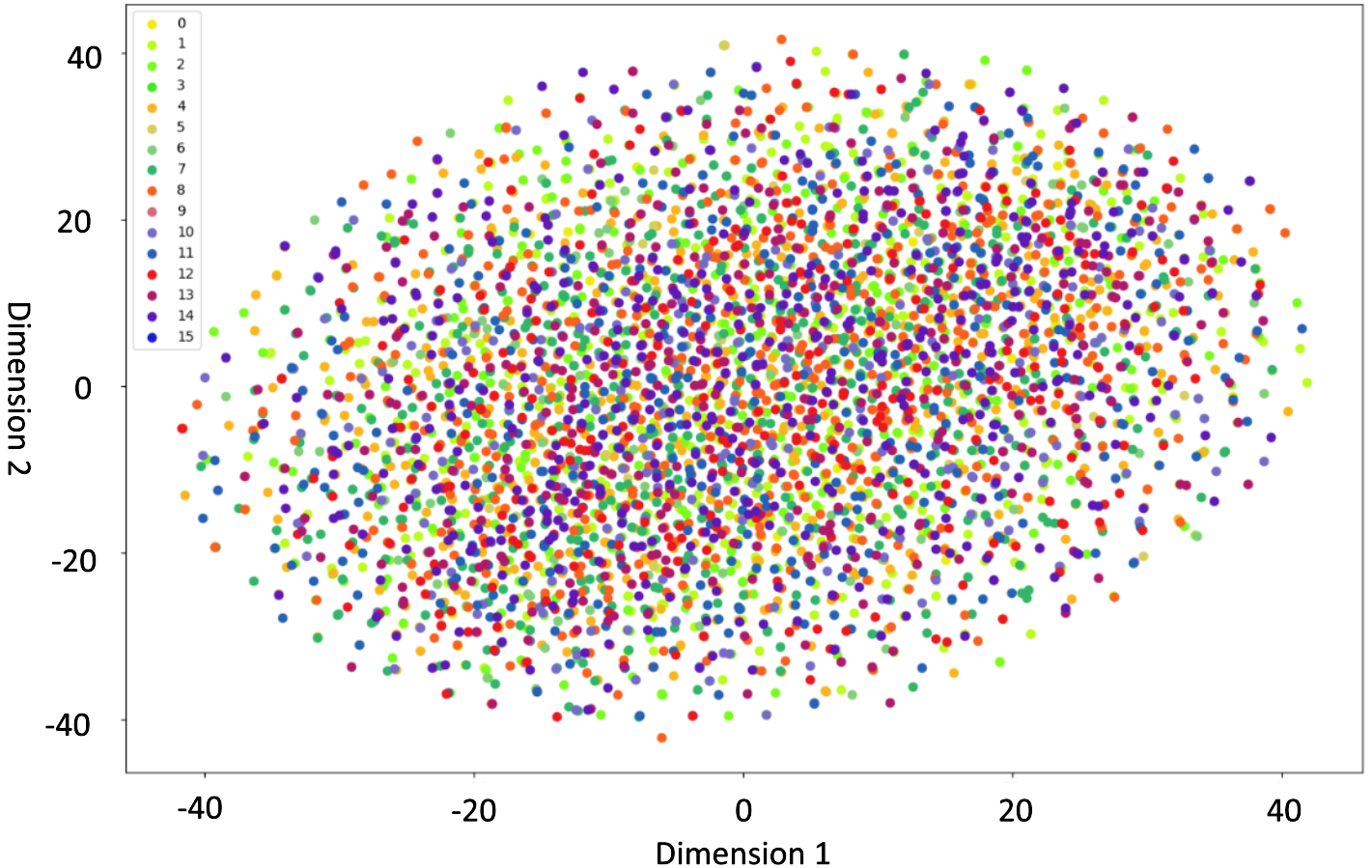}
          \includegraphics[height=0.99in]{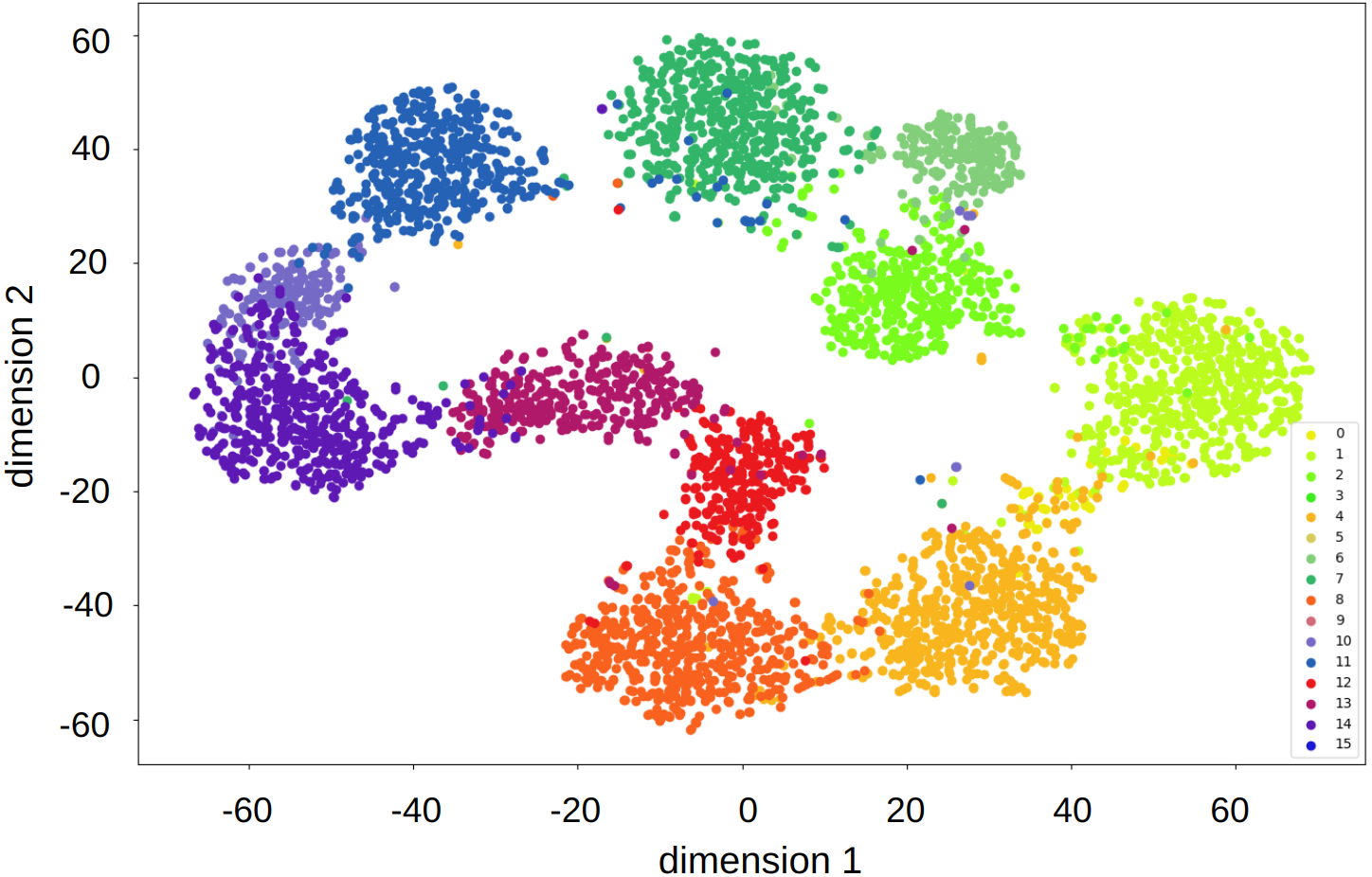}
      \includegraphics[height=0.99in]{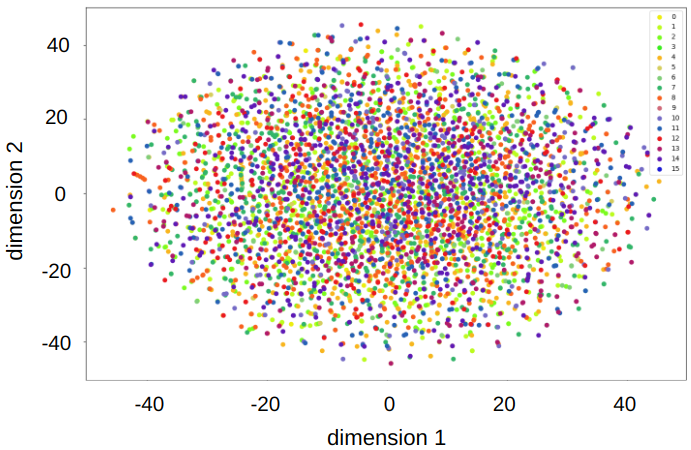}
\caption{T-SNE of 64-digit embeddings of ugVAE (left), 16-digit excitation portion (middle) and 48-digit inhibition portion (right) of gVAE16-v.}
\label{tsne_u_inhi}
\end{center}
\end{figure}

\begin{comment}
\begin{figure}[htbp]
\begin{center}
  \begin{minipage}{.32\textwidth}
\includegraphics[width=0.95\textwidth]{images/unguided_test.png}
\caption{T-SNE of 64-digit embeddings of ugVAE}
\label{tsne_u}
  \end{minipage}
  \hspace{3pt}
  \begin{minipage}{.65\textwidth}
    \centering
          \includegraphics[width=0.48\textwidth]{images/dark_207.png}
      \includegraphics[width=0.48\textwidth]{images/inhibition.png}
\caption{T-SNE of 16-digit excitation portion (left) and 48-digit inhibition portion (right) of gVAE16-v.}
\label{inhi}
  \end{minipage}
\end{center}
\vskip -0.2in
\end{figure}
\end{comment}

\begin{figure}
\centering
\includegraphics[width=\textwidth]{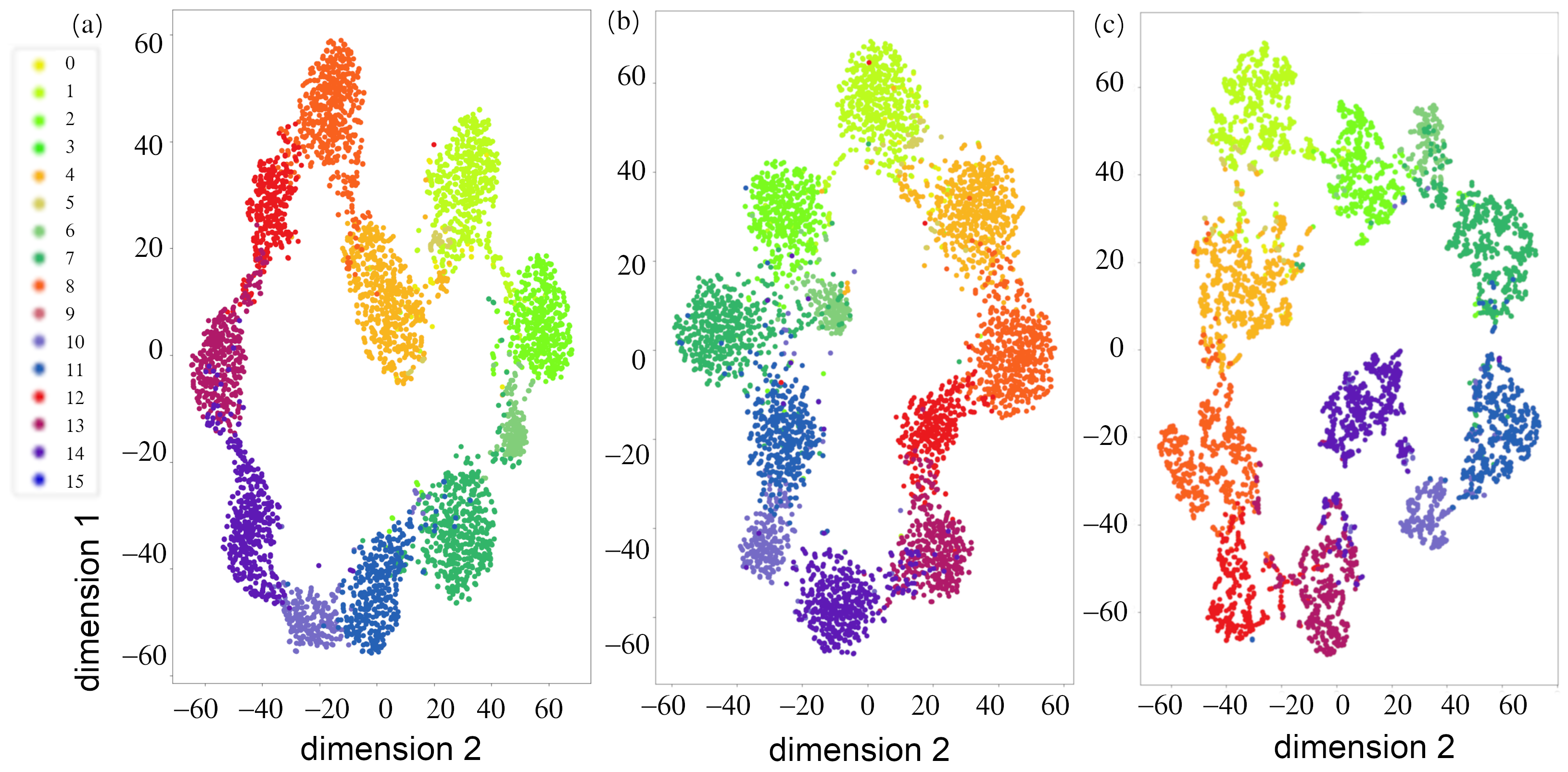}
\caption{Testing results of T-SNE visualization of excitation portion of latent codes, encoded by gVAE16 (a), gVAE8 (b), and gVAE4 (c).}
\label{tsne}
\end{figure}
Simultaneously, an excitation classifier takes the initial 16 latent variables to predict a location label while undergoing joint training with the encoder, in order to force the encoder to encode spatial features exclusively into the first 16 variables.
Conversely, an inhibition classifier operates on the remaining 48 variables in the latent space, engaged in adversarial training concerning two distinct target sets. 
One set still aligns with the target location features utilized by the excitation classifier. 
The other set, comprising a vector of length 16 with all values set to a constant number, denotes an absence of specific target correspondence.
This adversarial inhibition classifier is jointly trained with the encoder to prevent the remaining 48 latent variables from encoding information pertinent to location features, redirecting their focus towards other features such as illumination. Figure~\ref{guided} displays the structure of the hybrid guided VAE.

The excitation and inhibition losses are individually specified for every feature $\mathbf{m}$, outlined as follows for each one:

{\footnotesize
\begin{equation}
\mathcal{L}_{\text{Exc}}(z, m) = \max_{c_m}\left(\sum_{n=1}^{N} E_{q(z_m | x_n)} \log p_{cm}(y = y_m(x_n) | z_m)\right)
\end{equation}
\begin{equation}
\mathcal{L}_{\text{Inh}}(z, m) = \max_{k_m}\left(\sum_{n=1}^{N} E_{q(z_{\backslash\text{m}} | x_n)} \log p_{km}(y = y_m(x_n) | z_{\backslash \text{m}})\right)
\end{equation}
}
Where $c_{m}$ denotes the classifier making a prediction on the m-th feature in the guided space, and $k_{m}$ is a classifier making a prediction over m in the unguided space $z_{\backslash \text{m}}$ \cite{gesture}.
The overall loss function for hybrid guided VAE is:

{\footnotesize
\begin{equation}
\mathcal{L}_{\text{overall}} = \mathcal{L}_{\beta\text{VAE}} + \mathcal{L}_{\text{Exc}} + \mathcal{L}_{\text{Inh}}
% \mathcal{L}_{\text{overall}} = \mathcal{L}_{\text{unguided beta VAE}} + \mathcal{L}_{\text{Exc}} + \mathcal{L}_{\text{Inh}}
\end{equation}
}
%\tablename{\ref{table}} and Figure~\ref{guided} summarize the structure of hybrid guided VAE. 
%\input{table.txt}
%\input{train_algorithm.txt}
%The training procedures of this hybrid guided VAE is displayed in Algorithm \ref{algorithm}. 
%Figure~\ref{example} displays a few examples of input event frame samples and their corresponding latent representations encoded by the SNN encoder in hybrid guided VAE.
\begin{comment}
\begin{figure}[t]
\begin{center}
\includegraphics[width=0.8\textwidth]{images/to_latent.png}
\caption{A few examples of event frame samples (top) and corresponding latent codes  (bottom) encoded by hybrid guided VAE, reparametered to 8 $\times$ 8}
\label{example}
\end{center}
\vskip -0.2in
\end{figure}
\end{comment}
The value 16 for excitation variables is established to match the number of location classes defined in the arena. 
However, we also investigated the performance of this network with fewer excitation variables, 8 and 4. 

\begin{comment}
The hyperparameters of training hybrid guided VAE are recorded in Table~\ref{hyper}.

\begin{table}[htbp]
\caption{Hyperparameters of training hybrid guided VAE}
\label{hyper}
%\vskip 0.15in
\begin{center}
\begin{small}
\begin{sc}
\scalebox{0.86}{
\begin{tabular}{lcccr}
\toprule
Hyperparameter  & Value\\
\midrule
number of epochs     & up to 280\\
VAE beta  & 1.5\\
learning rate   & 3e-5\\
batch size & 16\\
latent dimension  & 64\\
excitation latent size  & 16, 8, 4\\
\bottomrule
\end{tabular}
}
\end{sc}
\end{small}
\end{center}
\vskip -0.1in
\end{table}
\end{comment}

\section{Results \& Evaluations}
\label{results}
The section analyses on the latent representations of the \ac{SNN} encoder. We visualize the latent codes through T-distributed stochastic neighbor embedding (T-SNE) \cite{tsne}, which displays the disentanglement of encoded latent code and the model's capability of generalization. Furthermore, we estimate the localization error of the robot by image retrieval.
\subsection{T-SNE visualization in latent space}
In the hybrid unguided VAE, all 64 variables equally contribute to reconstruction, with no distinction between disentangled and entangled features (Figure~\ref{tsne_u_inhi} left). In contrast, the hybrid guided VAE retains only 16, 8, or 4 excitatory variables for analysis.
They will be referred to as gVAE16, gVAE8, and gVAE4 respectively.
A T-SNE method evaluates this, where each dot represents a latent vector colored by the label in Figure~\ref{arena}. The unguided beta VAE shows disordered embeddings, implying it encodes many features without aligning to specific locations. However, the guided VAE better disentangles the latent space. While inhibition variables remain entangled (Figure~\ref{tsne_u_inhi} right), excitation variables are clearly separated (Figure~\ref{tsne}). For gVAE16, T-SNE reveals tight clusters for location classes, and clustering persists with gVAE8 and gVAE4, though clusters loosen (Figure~\ref{tsne} b,c). 
gVAE16, gVAE8, and gVAE4 achieved classification accuracies of 89\%, 89\%, and 83\% in testing (Table~\ref{results_table}). The guided VAE with 16 excitation variables trained and tested under varying illumination (gVAE16-v) achieved similar classification accuracy, with T-SNE embeddings showing slightly looser clustering (Figure~\ref{tsne_u_inhi} middle). This indicates the model disentangles locational from irrelevant features like illumination. 
Comparatively, an SNN model using weighted neuron assignment \cite{australia}, that reports to achieve performance comparable to NetVlad in other datasets, reached 86\% classification accuracy when finetuned and tested with our RGB dataset, highlighting the hybrid guided VAE's comparable performance in indoor VPR tasks.

\subsection{Cross-scene generalization}
\begin{figure}[t]
\includegraphics[height=.65in]{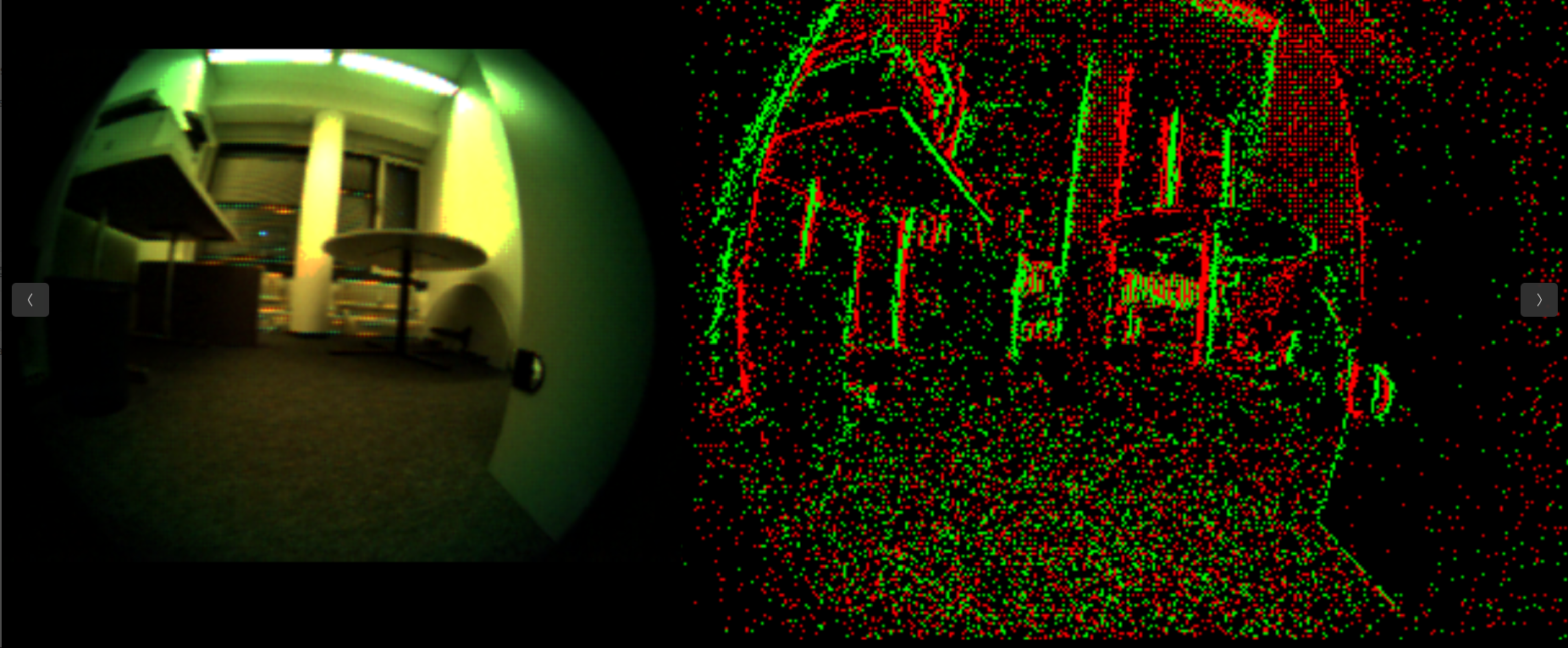}
\includegraphics[height=.65in]{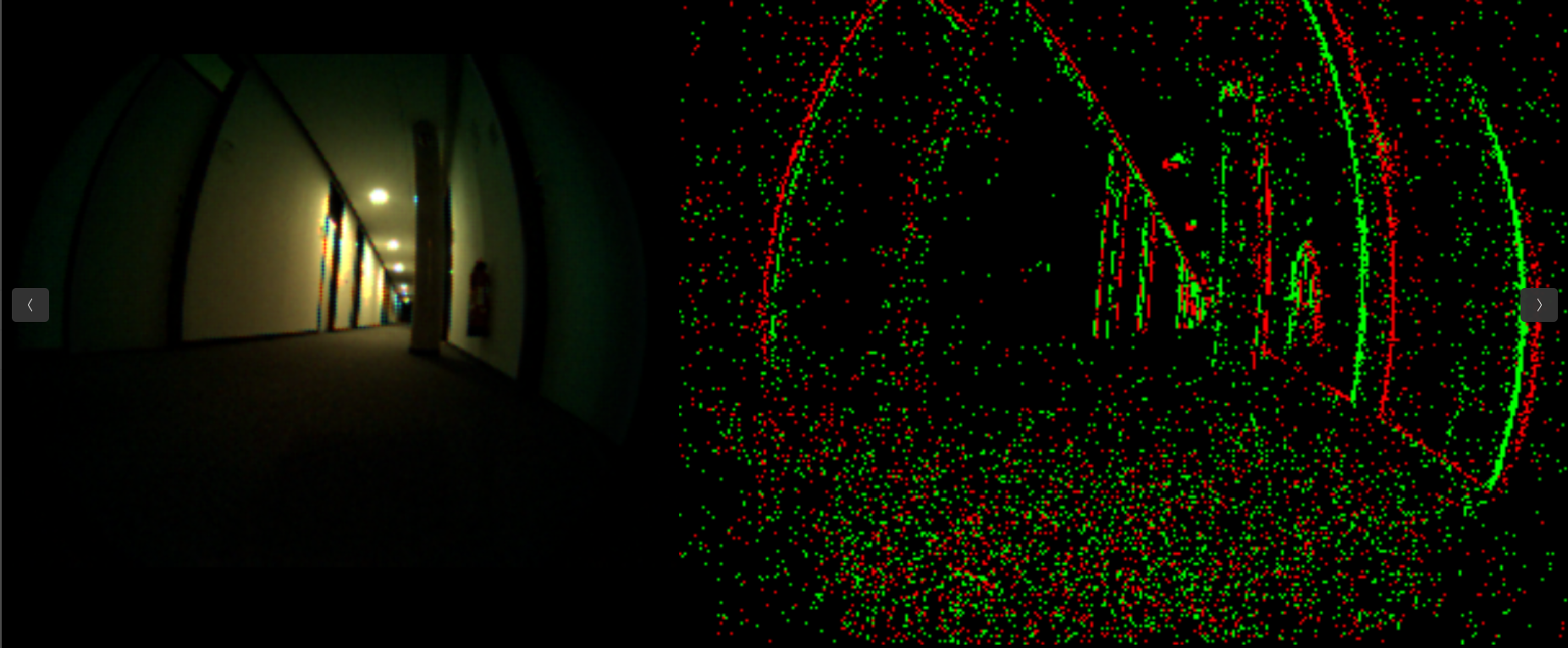}
\includegraphics[height=.65in]{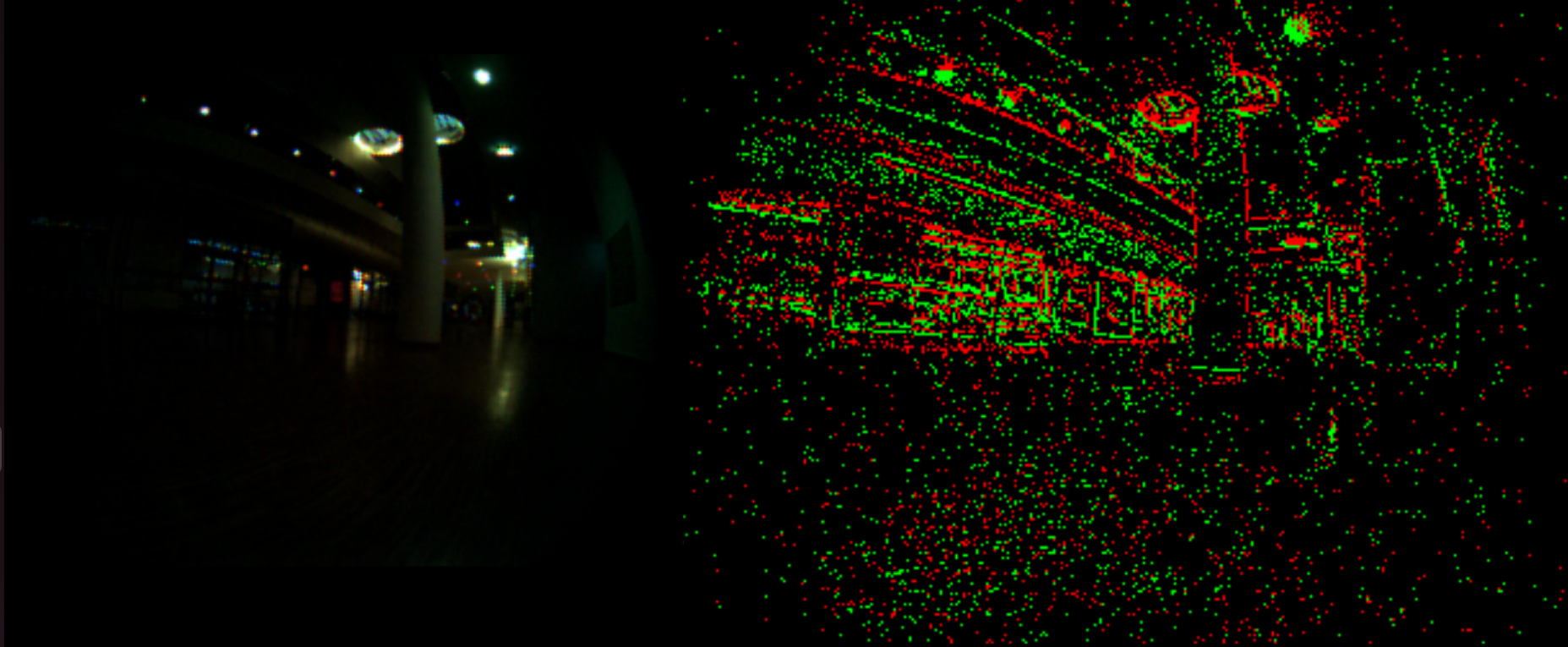}
\caption{New places recorded to evaluate generalization: printing room, passageway and hall (from left to right).}
\label{add}

\vspace{10pt}
  \centering
    \includegraphics[height=1.5in]{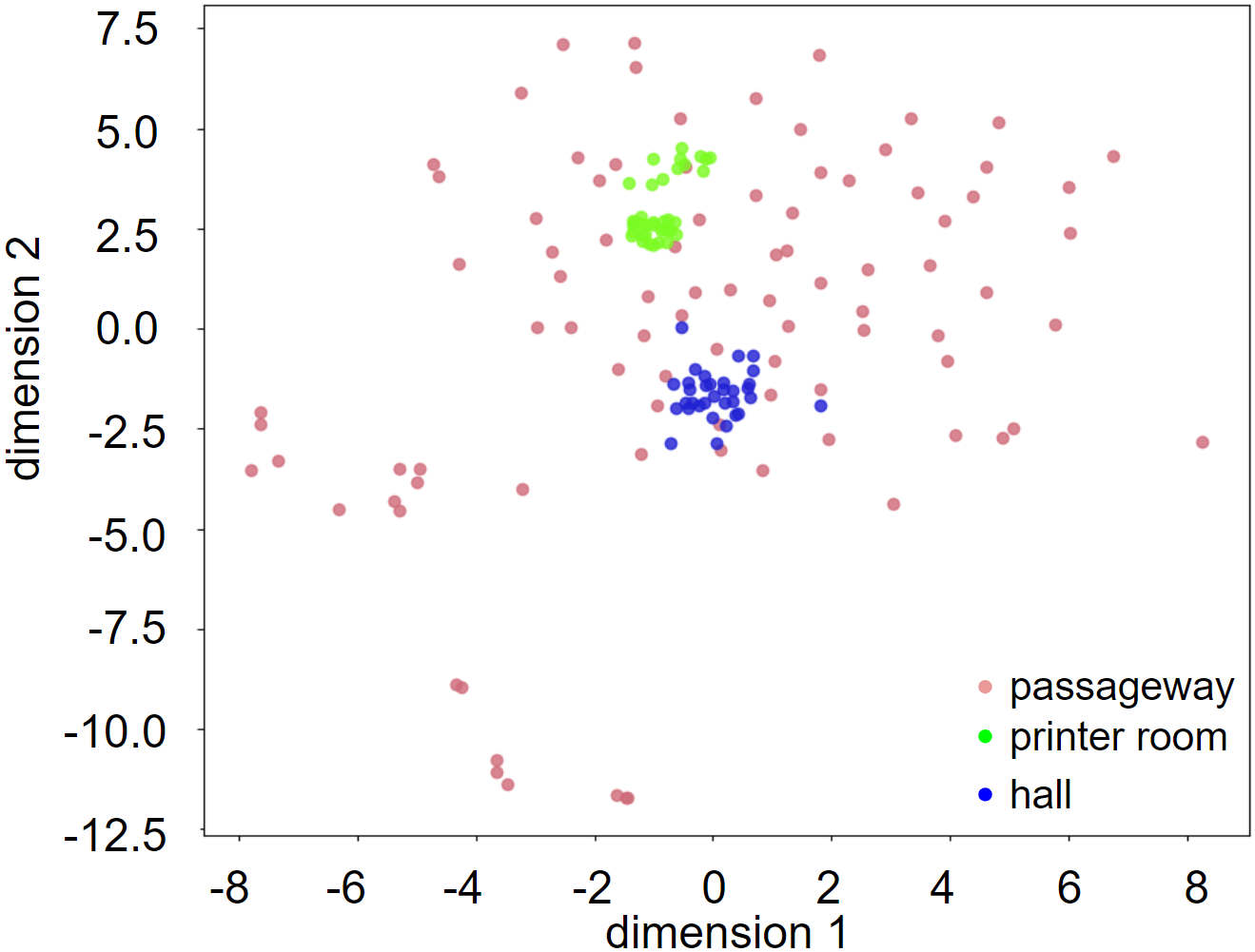}
    \includegraphics[height=1.55in]{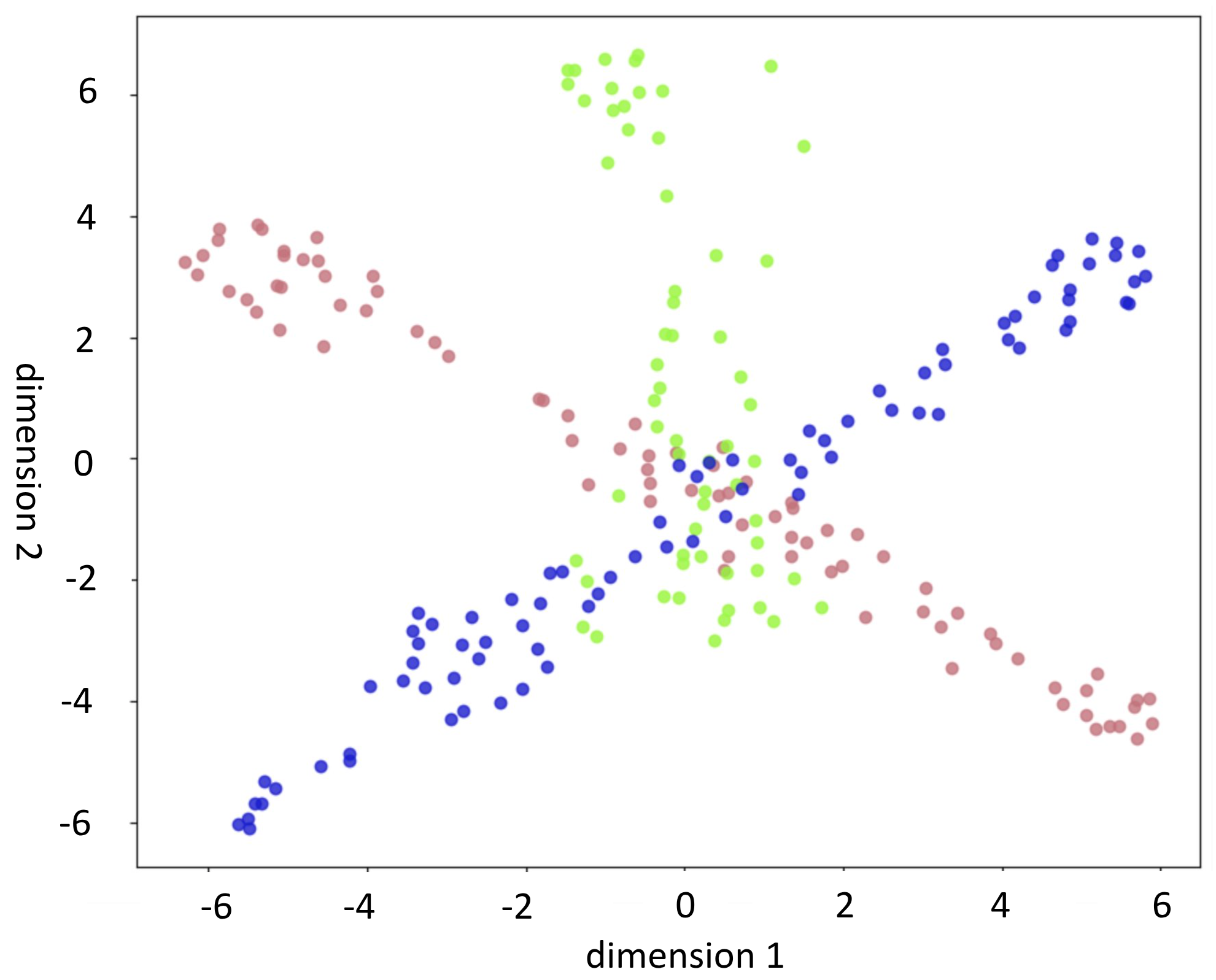}

    \includegraphics[height=1.3in]{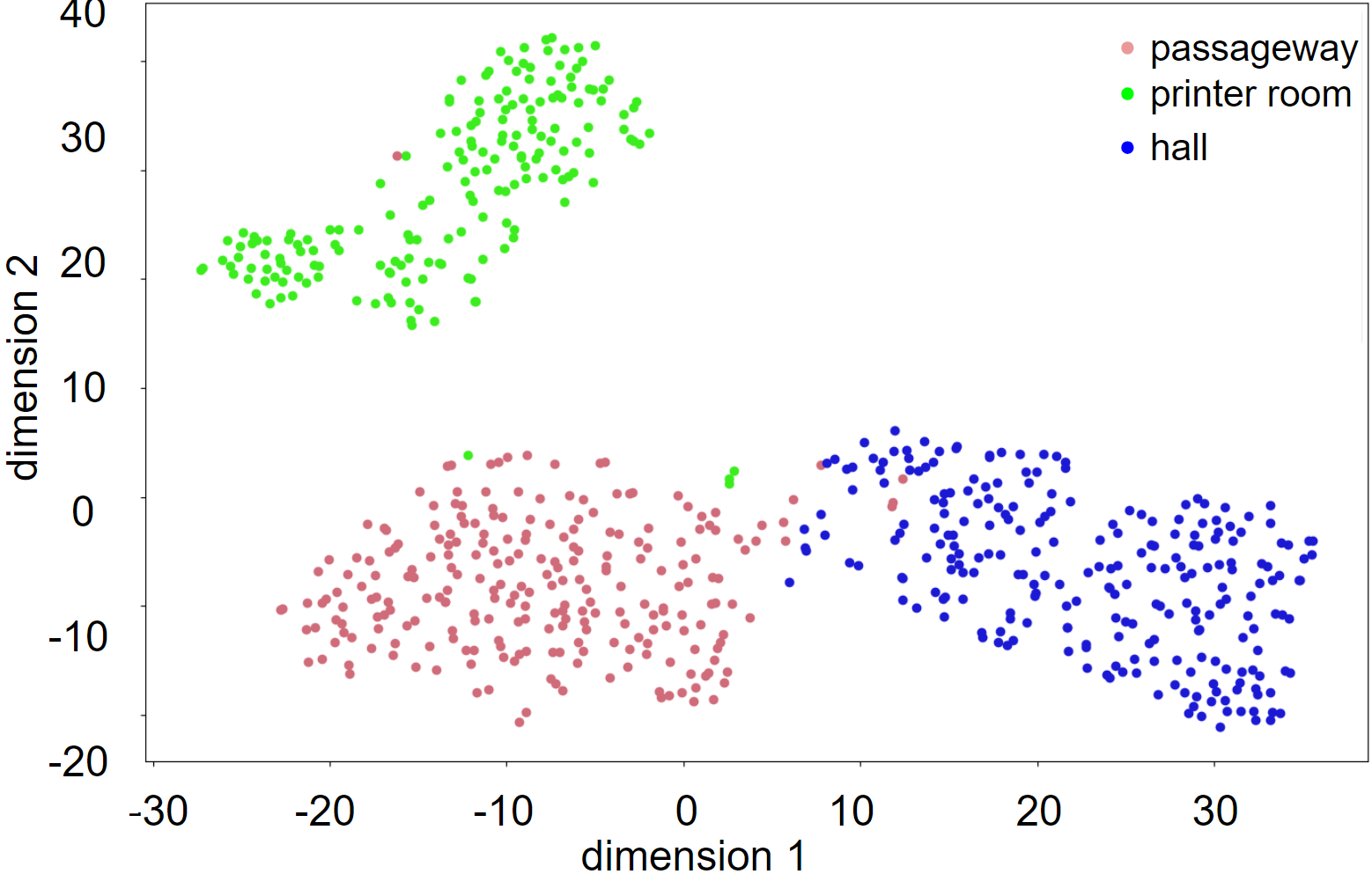}
    \includegraphics[height=1.43in]{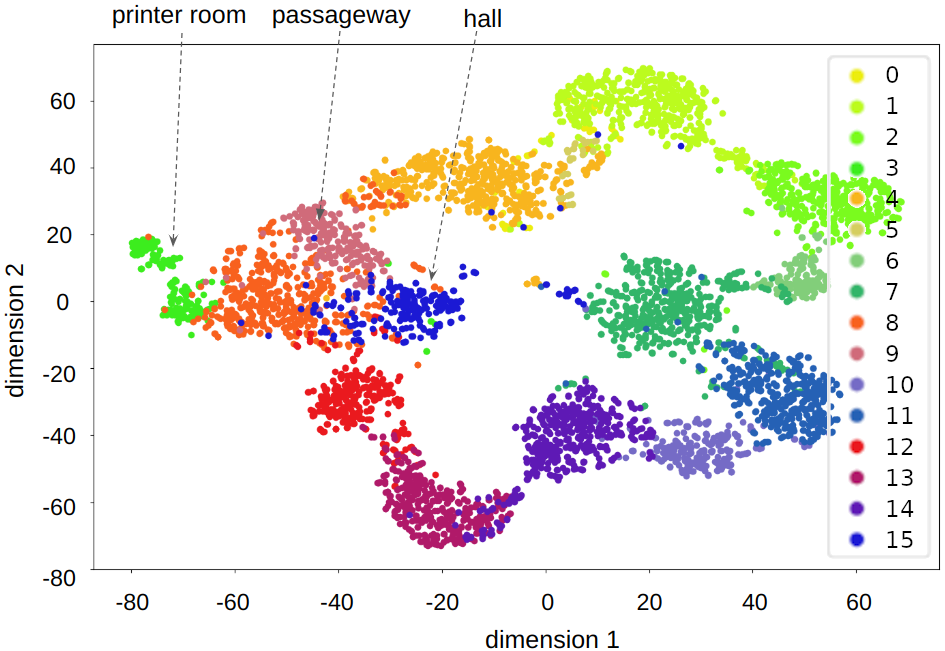}
\caption{Comparison of generalization. Top-left: Finetuned NetVlad (VGG16) with new places; top-right: finetuned NetVLAD attached to spiking VGG16 with new places; bottom-left: gVAE16-v with new places; bottom-right: gVAE16-v with three new places mixed in learnt places.}
\label{comparison}
\end{figure}

\begin{comment}
\begin{figure}[htbp]
  \centering
  \begin{minipage}{.3\textwidth}
    \centering
    \includegraphics[width=0.9\linewidth]{images/test_netvlad.png}
    \caption{T-SNE of pre-trained NetVlad (VGG16) with unseen dataset.}
    \label{testnetvlad}
  \end{minipage}
  \hspace{6pt}% Adjust the space as needed
  \begin{minipage}{.3\textwidth}
    \centering
    \includegraphics[width=0.9\linewidth]{images/spiking_vgg_netvlad.png}
    \caption{Generalization of spiking VGG16 trained with our dataset.}
    \label{testnetvlad}
  \end{minipage}
  \hspace{6pt}% Adjust the space as needed
  
  \begin{minipage}{.3\textwidth}
    \centering
    \includegraphics[width=1\linewidth]{images/dark_117_0.png}
\caption{T-SNE of gVAE16 (varying) with unseen dataset.}
\label{4new}
  \end{minipage}%
   \hspace{6pt}% Adjust the space as needed
  \begin{minipage}{.32\textwidth}
    \centering
    \includegraphics[width=1\linewidth]{images/3new_updated.png}
\caption{T-SNE of gVAE16 (varying) with three new places mixed in learnt places.}
\label{3new}
  \end{minipage}
\end{figure}
\end{comment}

%Zero-shot learning (ZSL) usually denotes a model's capacity to detect classes that are not part of its training set \cite{zeroshot}.
The generalization capability of a model refers to its ability to apply learned knowledge from training data to new, unseen data effectively.
We assess the trained encoder's capability to disentangle the feature vectors of unseen places, more specifically, to recognize new places not encountered during training.

Three additional datasets were studied including the robot maneuvering in a printing room, a passageway, and a high-ceiling hall, all of which were captured in slightly dimmer lighting compared to the dataset with normal illumination (Figure~\ref{add}).
Every new place provides approximately 50 new data points. 
The three new datasets are passed onto the frozen encoder trained by hybrid guided VAE with 16 excitatory variables on varying-illumination dataset (gVAE16-v), resulting in T-SNE embeddings depicted in Figure~\ref{comparison} (bottom-left). 
In the case of entirely new places, the embeddings exhibit clear disentanglement from each other and clustering, indicating gVAE16-v's proficiency in disentangling unknown places. 
Furthermore, the new places were tested together with the testing dataset from the learned arena. 
In the case of three new places mixed with 13 learnt places, the embeddings of new places cluster near the edge of class 8 (see Figure~\ref{arena}) without entangling with it (Figure~\ref{comparison} bottom-right). 
These results suggest that the trained encoder exhibits proficiency in cross-scene generalization. 
This capability empowers the robot to recognize and categorize new places when exploring an entirely novel environment without heavy continual learning.

%To compare the performance of this model with recent methods, the RGB frame recorded with the same camera is used for evaluating methods from literatures\cite{australia} and \cite{netvlad}.
%(with the aid of repository \cite{wild})

Comparatively, pre-trained NetVLAD (in the setting of backbone VGG16) finetuned with our RGB training dataset, succeeds in separating `printer room' and `hall', but not `passageway' when tested with the RGB dataset recorded in parallel with the three new places. NetVLAD was further tested by attaching to a spiking VGG16 and finetuned with our event dataset, was found to only partially generalizable to the three new scenes (Figure~\ref{comparison} top-left and top-right).
\subsection{Localize robot through image retrieval}
Many VPR approaches compare images sampled and stored during a previous exploration phase (reference image) to the current camera input (query image). 
Every reference image is assigned to a specific location. 
By finding the most similar reference image the robot can distinguish its approximate location.
In our evaluation, the location of the robot is estimated by computing the cosine similarity between two sequences, each of which consists of five successive excitation variable vectors.
We estimate the location of each query image by iterating through all reference samples of the training dataset.
Estimation based on a sequence instead of a single event frame is proven to eliminate accidental errors and decreases the probability that two places look the same\cite{seq}. 
The computation of cosine similarity follows:

\begin{equation}
\scalebox{0.95}{$
\text{C} = \frac{\sum_{i=1}^{5}\sum_{j=1}^{16}\text{SeqA}_{i,j}\text{SeqB}_{i,j}}
{\sqrt{\sum_{i=1}^{5}\sum_{j=1}^{16}\text{SeqA}_{i,j}^2}
 \sqrt{\sum_{i=1}^{5}\sum_{j=1}^{16}\text{SeqB}_{i,j}^2}}
$}
\end{equation}

Here, $\mathbf{SeqA}_{i,j}$ and $\mathbf{SeqB}_{i,j}$ represent the elements in vector sequences $\mathbf{SeqA}$ and $\mathbf{SeqB}$ respectively.
The coordinate %$\mathbf{(x, y)}$ 
of the middle element (I.e., the third out of five) in each sequence captured by motion capture system is regarded as the location of that sequence. 
%(visualized in Figure~\ref{sequ}). 
The coordinate of the reference sequence with the highest cosine similarity to the query sequence is regarded as the estimated location. 
The coordinate of the query event frame is regarded as the ground truth. 
Figure~\ref{his} displays the error distribution for the four models. Approximately 90.0\% of the samples exhibit an error of less than 0.5 meter in case of gVAE16. 
The small error is attributed to the fact that the excitation classifier is trained to differentiate locations at intervals of 1.5 meters. 
It is anticipated that the accuracy will further improve with more precise labeling.
gVAE8, gVAE4, and gVAE-v
%The hybrid guided VAE with 8 and 4 excitatory latent variables, and 16 excitatory variables in the cases of varying-illumination, 
achieve an error of less than 0.5 meter for  80.1\%, 77.5\% and 88.1\% samples, respectively.
These results have proven successful localization of the robot solely based on event camera input. 
Table~\ref{results_table} displays the evaluation metrics for different variations of hybrid VAEs, the SNN with weighted neuron assignment, and NetVlad.
\begin{table}[htbp]
\label{results_table}
\begin{center}
\caption{Evaluation metrics for hybrid VAEs and other methods, 
in terms of model size (number of parameters), TSNE results, classification accuracy,
localization accuracy, and cross-generalization.}

\scalebox{0.75}{
\begin{tabular}{lcccccr}
\toprule
model & param(test/train) & feature size & TSNE & classifier acc.(\%) & \textless 0.5m (\%) & cross-scene\\
\midrule
ugVAE  & 4.45M/5.65M & 64 & disordered & N.A. & N.A.& N.A.\\
gVAE16 & 4.45M/5.68M & 16 & clusters in loop & 89 & 90.0& N.A.\\
gVAE16-v & 4.45M/5.68M & 16 & clusters in loop & 89 & 88.1& generalizable\\
gVAE8 & 4.45M/5.68M & 8 & clusters in loop & 89 & 80.1& N.A.\\
gVAE4 & 4.45M/5.68M & 4 & clusters in loop & 83 & 77.5& N.A.\\
SNN\cite{australia}  & 0.6M/0.6M & 400 & N.A. & 86 & N.A.& N.A.\\
NetVlad\cite{netvlad} & 88M/88M & 64 & N.A. & N.A. & N.A.& partially generalizable\\
\bottomrule
\end{tabular}
}
\end{center}
\end{table}

%\footnotetext{The suffix “-v” indicates the model variant trained under varying illumination conditions.}

%source of result:
%gVAE4 77.5 epoch600 1outof3_lr

\subsection{Latent variable activity}
\begin{figure}[htbp]
  \centering
  \begin{minipage}{.42\textwidth}
    \centering
    \includegraphics[width=\linewidth]{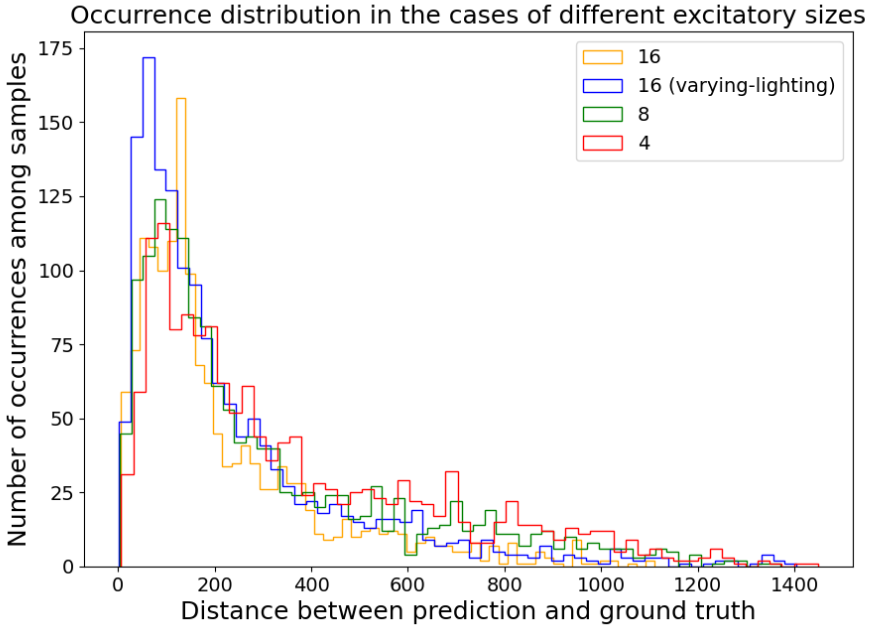}
    \caption{Distribution of localization errors of four models.}
    \label{his}
  \end{minipage}
  \begin{minipage}{.57\textwidth}
    \centering
    \includegraphics[width=1\linewidth]{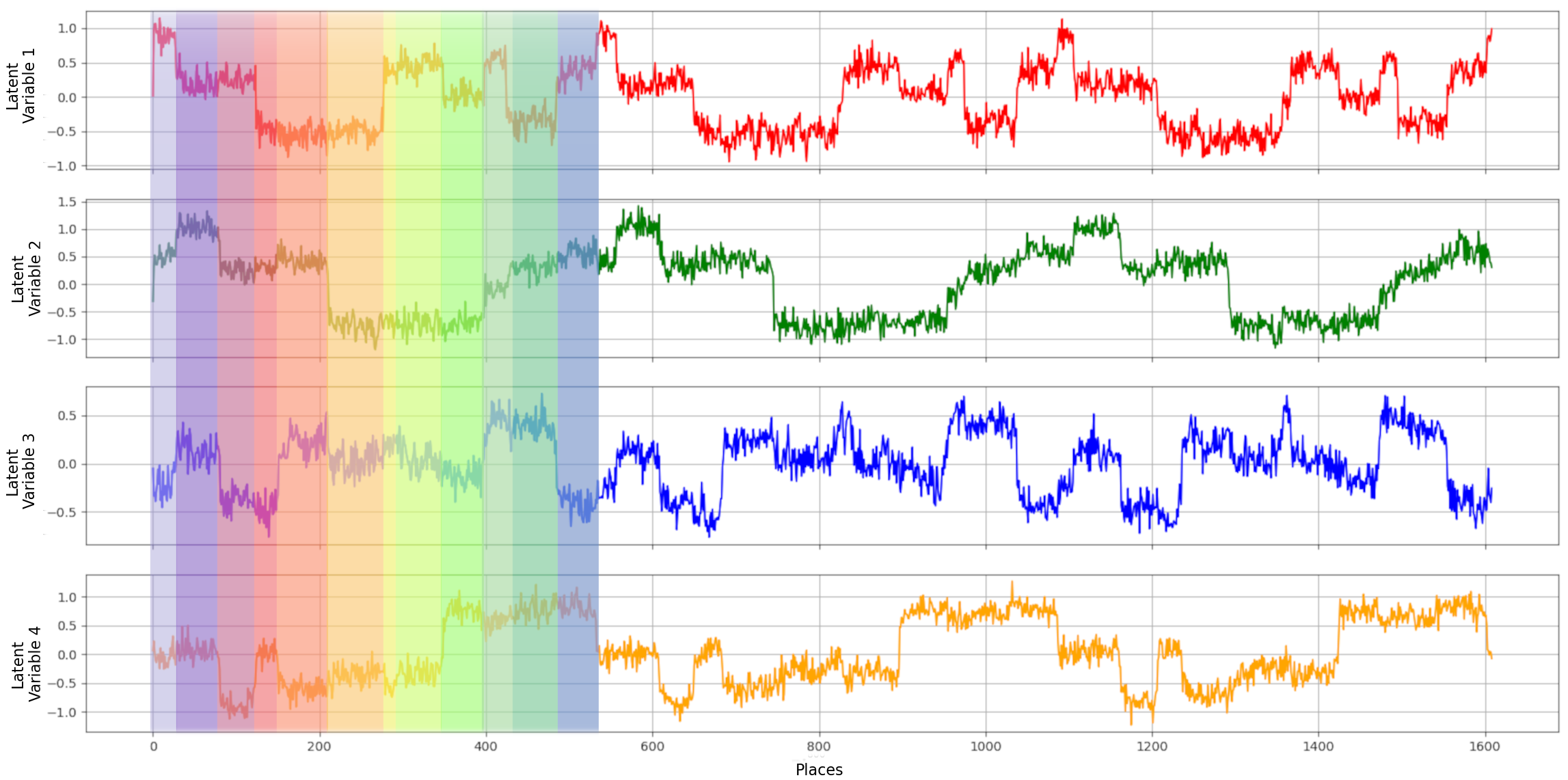}
\caption{Excitatory latent variable activities of gVAE4 over the trip of three rounds consisting of 1521 location samples. Vertical stripes mark the location class of the first round.}
\label{activity}
  \end{minipage}%
  %\hspace{6pt}% Adjust the space as needed
\end{figure}

To understand the internal representation of the model, we plotted the latent activity of 4 excitation variables of gVAE4 over three rounds of the agent's trip (Figure~\ref{activity}). The variables encode the agent's location in a multi-level fashion, with each variable displaying 4 or more distinct activation levels. Places are uniquely identified by a combination of these levels, with transitions between cells causing step-wise changes in some variables. With a minimum of four separable levels, these variables can encode up to $4^{4} = 256$ locations, enabling efficient storage of memory-intensive input images in a compact format. Emerging non-volatile analogue memristive devices are well-suited to store these analogue values, providing compact memory solutions \cite{analogmem2016, memreview2023}.
%The latent activities of 8 and 16 excitation variables over corresponding trips are displayed in Figures~\ref{tsne8} and ~\ref{tsne16} in Appendix.

\section{Conclusions \& Future Work}
\label{conclusions}
This project tackles the challenge of enhancing visual place recognition for mobile robot navigation in ambiguous indoor environments. We recorded a new indoor dataset for event-based visual place recognition. Based on this dataset we evaluated and improved the performance of a hybrid guided \ac{VAE} model.
The network can clearly disentangle different places while adjacent places remain close to each other (Figure \ref{tsne}). Investigations on the behaviour of the latent variable state indicate that the variables encode location in a step-wise multi-level code (Figure \ref{activity}). Such a code enables the model to easily represent a number of places greater than the number of excited variables in a latent vector. This very efficient compression method is leading to little memory utilization so that a large number of different places can be stored. Emerging devices such as memristive devices are a good fit for this kind of memory utilization due to their capability of analogue memory formation and their non-volatile behaviour \cite{analogmem2016}. 
When computing the cosine similarity between input image and previously visited reference places, the algorithm can distinguish the location of the robot with a localization error of less than 0.5 meters in 90.0, 80.1 and 77.5 percent of the cases for excitation variable sizes of 16, 8 and 4 respectively. 
Since the cells of the different place classes are 1.5m $\times$ 1m large the precision of the localization algorithm should be at the resolution of half the cell size. We improved this resolution by using a sequence of 5 successive latent vectors. These sequences can capture transitions from one class to another which are much higher spatial resolution than the cells themselves. Hence, even longer sequences of latent vectors might allow for an even higher localization precision as done by Milford in \cite{howlong2013}. Furthermore, we can increase the localization accuracy by increasing the number of labeled cells.
Besides the localization accuracy we tested the classification accuracy reaching 90, 89 and 83 percent with 16, 8 and 4 excited variables respectively. The state of the art \ac{SNN} model from Hussaini et al. \cite{australia} reaches an accuracy of 86 percent on our dataset. Therefore, gVAE16 performs the best on our new indoor \ac{VPR} dataset. When comparing \cite{australia} and our model it can be noted that Hussaini et al. only requires 600k parameters while our model consists of around 4.45M parameters. However, our model provides a much more efficient latent space compression by storing a high number of locations in only four latent variables in contrast to 400 variables for 100 places. To our knowledge, \cite{australia} and our approach are the only two existing \ac{SNN} approaches for \ac{VPR}. Hussaini et al. beats NetVlad \cite{netvlad} on several large datasets such as Nordland, Oxford Robot Car, SPEDTest, Synthia and St Lucia and is therefore a good baseline comparison. \\
Besides its capability to disentangle different places and localize itself our model can also distinguish between known and unknown places. When presented with new places the model generated new clusters in the T-SNE representation. 
Also, the model can successfully disentangle new places and outperformed finetuned NetVlad attached to either VGG or spiking VGG when tested on our new dataset. These findings illustrate that the model learns the general features of location and can therefore clearly distinguish between unseen places. Such a model has the capability to be used in unknown environments without requiring any expensive continual learning. The generalization capability of the model will be further investigated on a larger dataset of unseen environments. Furthermore, the model shows strong robustness against environmental changes. When trained with varying lighting conditions the model is able to correctly localize itself with a localization error of less than 0.5 meters in 88.1 percent of the cases and a classification accuracy of 89 percent.
With only 4.45M parameters and 917K neurons the model has the potential to fit on existing low-power low-latency neuromorphic hardware. With a few small changes the spiking encoder model can directly be ported onto such a processor, which has already been shown by \cite{beta}. Data from an event camera can directly be streamed onto the neuromorphic processor leading to a hardware implementation well suited for compact and energy efficient robotic systems.
Thus, our guided \ac{VAE} model enables \ac{VPR} in known and unknown environments implemented on low-power low-latency neuromorphic hardware. This work uncovers the potential of hybrid guided \acp{VAE} for \ac{VPR}, paving the way towards a new generation of robotics based on emerging brain-inspired processors, sensors and devices.

\bibliographystyle{IEEEtran}
\bibliography{main}





% LNCS Homepage, \url{http://www.springer.com/lncs}. Last accessed 4
% \end{thebibliography}
\end{document}